\pdfoutput=1
\documentclass[11pt,a4paper]{article}
\usepackage[hyperref]{acl2020}
\usepackage{times}
\usepackage{latexsym}
\usepackage{amsmath}
\usepackage{amsfonts}
\usepackage{amssymb}
\usepackage{graphicx}
\usepackage{CJKutf8}
\usepackage{booktabs}
\usepackage{ulem}

\usepackage{microtype}

\aclfinalcopy 

\title{CPM: A Large-scale Generative Chinese Pre-trained Language Model}

\author{Zhengyan Zhang, Xu Han, Hao Zhou, Pei Ke, Yuxian Gu, Deming Ye, \\
\textbf{Yujia Qin, Yusheng Su, Haozhe Ji, Jian Guan, Fanchao Qi, Xiaozhi Wang, Yanan Zheng,} \\
\textbf{Guoyang Zeng, Huanqi  Cao, Shengqi Chen, Daixuan Li, Zhenbo Sun,} \\
\textbf{Zhiyuan Liu$^\dagger$, Minlie Huang$^\dagger$, Wentao Han, Jie Tang, Juanzi Li, Xiaoyan Zhu, Maosong Sun} \\
Department of Computer Science and Technology, Tsinghua University \& BAAI}

\date{}

\begin{document}
\maketitle

\begin{CJK*}{UTF8}{gbsn}

\begin{abstract}
Pre-trained Language Models (PLMs) have proven to be beneficial for various downstream NLP tasks. Recently, GPT-3, with 175 billion parameters and 570GB training data, drew a lot of attention due to the capacity of few-shot (even zero-shot) learning. However, applying GPT-3 to address Chinese NLP tasks is still challenging, as the training corpus of GPT-3 is primarily English, and the parameters are not publicly available. In this technical report, we release the Chinese Pre-trained Language Model (CPM) with generative pre-training on large-scale Chinese training data. To the best of our knowledge, CPM, with 2.6 billion parameters and 100GB Chinese training data, is the largest Chinese pre-trained language model, which could facilitate several downstream Chinese NLP tasks, such as conversation, essay generation, cloze test, and language understanding. Extensive experiments demonstrate that CPM achieves strong performance on many NLP tasks in the settings of few-shot (even zero-shot) learning. The code and parameters are available at \href{https://github.com/TsinghuaAI/CPM-Generate}{https://github.com/TsinghuaAI/CPM-Generate}. 
\end{abstract}

\section{Introduction}

{\let\thefootnote\relax\footnotetext{$^\dagger$ Corresponding authors: Z. Liu (liuzy@tsinghua.edu.cn) and M. Huang (aihuang@tsinghua.edu.cn)}}

Pre-trained Language Models (PLMs)~\cite{peters2018deep,radford2018improving,devlin2018bert,brown2020language} have been developed for a variety of tasks in Natural Language Processing (NLP), as they can learn rich language knowledge from large-scale corpora, which is beneficial for downstream tasks. ELMo~\cite{peters2018deep} first introduces bidirectional language models to learn contextual word vectors via large-scale pre-training. GPT~\cite{radford2018improving} applies generative pre-training to a Transformer-based language model~\cite{vaswani2017attention}, which improves natural language understanding on a wide range of benchmarks. BERT~\cite{devlin2018bert} is proposed to pre-train deep bidirectional representations on unlabeled texts by jointly conditioning on both left and right contexts. RoBERTa~\cite{liu2019roberta} and ALBERT~\cite{lan2019albert} enhance BERT~\cite{devlin2018bert} by dynamic masking, parameter sharing and modifying pre-training tasks. ERNIE~\cite{zhang2019ernie}, KEPLER~\cite{wang2019kepler} and SentiLARE~\cite{ke-etal-2020-sentilare} introduce external knowledge to language representation learning by auxiliary pre-training tasks. 

Among these PLMs, GPT-3~\cite{brown2020language}, with 175 billion parameters and 570GB training data, has been the center of attention and proven to be effective in various few-shot (even zero-shot) NLP tasks. The powerful text generation capability of GPT-3 makes it available to diverse applications, such as question answering, summarization, conversation, computing basic arithmetic, and generating kinds of text, including essay, fiction, code, spreadsheets, etc. However, incorporating GPT-3 to address Chinese NLP tasks is still challenging, as the training corpus of GPT-3 is primarily English (93\% by word counting as reported by \citet{brown2020language}), and the parameters are not publicly available. Although there are some previous works providing powerful Chinese pre-trained language models~\cite{cui-etal-2020-revisiting,xu2020clue,wei2019nezha,sun2019ernie,cui2019pre}, their capabilities are limited due to the model size. Hence, how to pre-train a large-scale Chinese language model needs more exploration, such as the construction of Chinese vocabulary and the design of the training strategy.

In this technical report, we release the \textbf{C}hinese \textbf{P}re-trained Language \textbf{M}odel (\textbf{CPM}) with generative pre-training on large-scale Chinese corpora. CPM is a Transformer-based autoregressive language model, with 2.6 billion parameters and 100GB Chinese training data. To the best of our knowledge, CPM is the largest Chinese pre-trained language model, which could facilitate downstream Chinese NLP tasks, such as conversation, essay generation, cloze test, and language understanding.
Experiments on various Chinese NLP tasks demonstrate that CPM achieves strong performance on many NLP tasks in the few-shot (even zero-shot) settings. With the increase of parameters, CPM performs better on most datasets, indicating that larger models are more proficient at language generation and language understanding.

The main contributions of this technical report can be summarized as follows:
\begin{itemize}
    \item We release a Chinese autoregressive language model with generative pre-training, called CPM, which has 2.6 billion parameters.
    
    \item We construct a new sub-word vocabulary based on the word segmented corpus to adapt for Chinese corpora and increase the batch size to $3,072$ for more stable model training.
    
    \item Extensive experiments demonstrate that CPM achieves strong performance on many NLP tasks in the few-shot (even zero-shot) settings.
    
\end{itemize}

\section{Our Approach}

\subsection{Chinese PLM}

\begin{table}[t]
\centering
\resizebox{\linewidth}{!}{
\begin{tabular}{lrrrrr}
\hline
 & $n_{\text{param}}$ & $n_{\text{layers}}$ & $d_{\text{model}}$ & $n_{\text{heads}}$ & $d_{\text{head}}$ \\ \hline
CPM-Small  &     109M     & 12        & 768      & 12       & 64      \\ 
CPM-Medium &     334M     & 24        & 1,024     & 16       & 64      \\ 
CPM-Large  &    2.6B      & 32        & 2,560     & 32       & 80      \\ \hline
\end{tabular}
}
\caption{Model sizes. $n_{\text{param}}$ is the number of parameters. $n_{\text{layers}}$ is the number of layers. $d_{\text{model}}$ is the dimension of hidden states, which is consistent in each layer. $n_{\text{heads}}$ is the number of attention heads in each layer. $d_{\text{head}}$ is the dimension of each attention head.}
\label{tab:model-size}
\end{table}

Our current model is a left-to-right Transformer decoder, which is similar to the model architecture of GPT~\cite{radford2019language}. We pre-train three models with different sizes, as shown in Table~\ref{tab:model-size}. In order to adapt CPM to Chinese corpora, we build a new sub-word vocabulary and adjust the training batch size. 

\textbf{Vocabulary Construction:} Previous works on Chinese pre-trained models usually adopt the sub-word vocabulary of BERT-Chinese~\cite{devlin2018bert}, which would split the input text to a character-level sequence. However, Chinese words usually contain several characters, and some important semantic meanings of words would be lost in the character-level sequence. To solve this problem, we construct a new sub-word vocabulary, containing both words and characters. For example, some common words would be added to the vocabulary. 

\textbf{Training Strategy:} Since the sparseness of word distributions of Chinese is more serious than that of English, we adopt a large batch size to make the model training more stable. Compared to the batch size (1 million tokens) used in GPT-3 2.7B~\cite{brown2020language}, our batch size (3 million tokens) is two times larger. For the largest model, which cannot be stored in a single GPU during training, we partition the model across GPUs along the width dimension to make the large-scale training available and reduce data-transfer among nodes.

\subsection{Data Processing}

Specifically, we construct a new sub-word vocabulary based on the word segmented corpus using \textit{unigram language model}~\cite{DBLP:conf/emnlp/KudoR18}. Meanwhile, considering that the word segmentation introduces extra splitters between words, we set a special token as the splitter to make the sub-word process reversible. In contrast, the tokenizer of BERT-Chinese is irreversible because it will insert extra spaces between Chinese characters and treat the extra spaces as the same as the original spaces in the text. 

\begin{table}[t]
\centering
\resizebox{\linewidth}{!}{\begin{tabular}{lccccc}
\hline
Data Source & Encyclopedia & Webpage & Story & News & Dialog \\ \midrule
Size  &     $\sim$ 40GB     & $\sim$ 39GB        & $\sim$ 10GB      & $\sim$ 10GB       & $\sim$ 1GB      \\  \hline
\end{tabular}}
\caption{Details of training data.}
\label{tab:training-data}
\end{table}

 We collect different kinds of texts in our pre-training, including encyclopedia, news, novels, and Q\&A. The details of our training data are shown in Table~\ref{tab:training-data}. Since the input sequence length is usually larger than that of a single document, we concatenate different documents together by adding ``end of document'' token after each document to make full use of the input length.

\subsection{Pre-training Details}

Based on the hyper-parameter searching on the learning rate and batch size, we set the learning rate as $1.5\times10^{-4}$ and the batch size as $3,072$, which makes the model training more stable. In the first version, we still adopt the dense attention and the max sequence length is $1,024$. We will implement sparse attention in the future. We pre-train our model for $20,000$ steps, and the first $5,000$ steps are for warm-up. The optimizer is Adam \cite{kingma15adam}. It takes two weeks to train our largest model using $64$ NVIDIA V100.


\section{Experiments}

\subsection{Text Classification}

\textbf{Dataset}: We use TouTiao News Titles Classification (TNEWS), IFLYTEK app description classification (IFLYTEK), and Original Chinese NLI (OCNLI) as our benchmark datasets for text classification~\cite{xu2020clue,hu2020ocnli}. Since we aim to evaluate the zero-shot ability of CPM on text classification tasks, we directly use the validation sets of these three datasets without any training 
instance. The amount of the validation set of TNEWS / IFLYTEK / OCNLI is 10K / 2.6K / 3K. Note that, we exclude the instances with the label ``-'' in OCNLI.

\textbf{Implementation Details}: We calculate the perplexity of each candidate sentence-label pair and treat the pair having the lowest perplexity as the prediction. The templates of these three tasks are formulated by
\begin{equation*}
\begin{aligned}
    \text{TNEWS:} & \ \ \text{这是关于}\uline{L}\text{的文章：}\uline{P} \\
    & \ \ (\text{This passage is about }\uline{L}\text{: }\uline{P}), \\
    \text{IFLYTEK:} & \ \ \text{这是关于}\uline{L}\text{的应用程序：}\uline{P} \\
    & \ \ (\text{This application is about }\uline{L}\text{: }\uline{P}), \\
    \text{OCNLI:} & \ \ \uline{S_1}\text{？对，}\uline{S_2} \ (\uline{S_1}\text{? Yes, }\uline{S_2}),\\
    & \ \ \uline{S_1}\text{？错，}\uline{S_2} \ (\uline{S_1}\text{? No, }\uline{S_2}),\\
    & \ \ \uline{S_1}\text{？也许，}\uline{S_2} \ (\uline{S_1}\text{? Maybe, }\uline{S_2}),\\
\end{aligned}
\end{equation*}
where $L$ is the label name, $P$ is the input text, $S_1$ and $S_2$ are the premise and hypothesis.

Since TNEWS and IFLYTEK have more than 10 kinds of labels, we adopt a simpler validation setting, which randomly samples 3 false labels for each instance and performing 4-class classification for better efficiency. To make it more stable, we repeat it three times and report the averaged results. For OCNLI, which only has 3 kinds of labels, we hold the original validation set. However, the validation set of OCNLI is unbalanced, where the amount of ``entailment'' / ``neutral'' / ``contradiction'' is 947 / 1103 / 900. If the model only predicts the label ``neutral'', the accuracy is about $0.374$.

\begin{table}[t]
\centering
\resizebox{\linewidth}{!}{\begin{tabular}{lccc}
\hline
 & TNEWS & IFLYTEK & OCNLI \\ \hline
CPM-Small  & 0.626 & 0.584   & 0.378 \\ 
CPM-Medium & 0.618 & 0.635   & 0.379 \\ 
CPM-Large  & \textbf{0.703} & \textbf{0.708}   & \textbf{0.442} \\ \hline
\end{tabular}}
\caption{Zero-shot performance on text classification tasks (accuracy). Random prediction would have $0.25$ on TNEWS and IFLYTEK, $0.33$ on OCNLI.}
\label{tab:txt-cls}
\end{table}

\textbf{Results}: As shown in Table~\ref{tab:txt-cls}, CPM-large achieves promising results on these classification datasets without any training samples. Compared to random prediction, the knowledge learned from pre-training significantly improves the performance. Although the medium model is three times as large as the small model, the performances on TNEWS and OCNLI are very close. However, CPM-Large significantly outperforms these two smaller models on all three datasets. It indicates that the magic of the model size is not linear and would happen when the model size exceeds a specific boundary. Besides, the results of CPM-small and CPM-medium on OCNLI are close to that of the strategy only predicting the label ``neutral''. It suggests that natural language inference is harder than other downstream tasks in the setting of zero-shot learning, which is consistent with the observation in \citeauthor{brown2020language}.

\begin{table}[t]
    \centering
    \begin{tabular}{lcc}
        \hline
                & Supervised & Unsupervised \\
        \hline
        CPM-Small &   0.657       &   0.433           \\
        CPM-Medium &  0.695          &   0.524             \\
        CPM-Large &   \textbf{0.804}       &   \textbf{0.685}         \\
        \hline
    \end{tabular}
    \caption{Results on ChID dataset in the supervised and unsupervised settings. The random prediction would have 0.10 in the unsupervised setting.}
    \label{tab:chid}
\end{table}

\begin{table*} [t]
\centering
\small
\setlength{\tabcolsep}{1.5mm}{
\begin{tabular}{lccccc}
\hline
& Average & Extrema & Greedy & Dist-1 & Dist-2 \\
\hline
\textit{Few-shot (Unsupervised)} &  &  &  &  & \\
CDial-GPT & 0.899 & 0.797 & 0.810 & 1,963 / \textbf{0.011} & 20,814 / 0.126 \\
CPM-Large  & \textbf{0.928} & \textbf{0.805} & \textbf{0.815} & \textbf{3,229} / 0.007 & \textbf{68,008} / \textbf{0.154}  \\
\hline
\textit{Supervised} &  &  &  &  & \\
CDial-GPT & 0.933 & \textbf{0.814} & \textbf{0.826} & 2,468 / 0.008 & 35,634 / 0.127 \\
CPM-Large & \textbf{0.934} & 0.810 & 0.819 & \textbf{3,352} / \textbf{0.011} & \textbf{67,310} / \textbf{0.233} \\
\hline
\end{tabular}}
\caption{Results on STC dataset in the few-shot and supervised settings.}
\label{tab:stcresult}
\end{table*}

\subsection{Chinese Idiom Cloze}

\textbf{Dataset: } We use the Chinese IDiom cloze test dataset (ChID)~\cite{zheng-etal-2019-chid} as our benchmark dataset. Each passage in the dataset may contain multiple blanks. For each blank, there are 10 candidate idioms with 1 golden truth. Some of the false candidates are similar to the answer in meanings. The amount of training / validation / test set is 520K / 20K / 20K.

\textbf{Implementation Details: } For the supervised setting, we use a template to convert the passage and the candidates to a natural language question. Given the passage $P$ and 10 candidate idioms $I_1, I_2, ..., I_{10}$, the template can be formulated as

\begin{equation*}
\begin{aligned}
    \text{选项1:} \ \uline{I_1} \ ... \ \text{选项10:} \ \uline{I_{10}} \ \uline{P} \ & \text{答案是:}\uline{L} \\
    (\text{Option 1:} \ \uline{I_1} \ ... \ \text{Option 10:} \ \uline{I_{10}} \ \uline{P} \ & \text{Answer:}\uline{L}). 
\end{aligned}
\end{equation*}

Then, we train the model to predict the answer $L$. Note that if there exists more than one idiom in a passage, we predict each one independently. Specifically, When we are predicting one idiom, we leave its blank in the passage and remove the blanks of other idioms from the passage. 

For the unsupervised setting, we fill the candidate idioms into the blank to form a group of complete passages. We also consider each idiom blank individually if there are multiple blanks in a passage. For each blank, we can get 10 passages corresponding to the 10 candidate idioms. Then we calculate the perplexity of each passage and treat the one with the lowest perplexity as the prediction. 

\textbf{Results: } The results are shown in Table \ref{tab:chid}. We report the accuracy on the test set of each model. For the fully supervised setting, we can see that CPM can be fine-tuned for the specific input template, solving multiple-choice tasks by uni-direction auto-regressive language modeling. In our experiments, we didn't take much time to design the input template for this task, and thus there might exist better templates that can help the model to show its full ability. We will leave this part as future work. For the unsupervised setting, we can see that CPM produces promising results. The unsupervised result of CPM-Large even outperforms the result of CPM-Small and is comparable to CPM-Medium in the supervised setting, reflecting the strong power of CPM in Chinese language modeling.


\begin{table} [t]
\centering
\footnotesize
\setlength{\tabcolsep}{0.8mm}{
\begin{tabular}{ll}
\hline
Post: & 现在高中数学竞赛这么凶残了！ \\
& How crazy the math competition of high \\
& school is! \\
\hline
CDial-GPT: & 你还真信？ \\
(Few-shot) & Do you really believe it? \\
CDial-GPT: & 数学让我们学会了很多，好好学，明天 \\
(Supervised) & 回家 \\
& Math lets us learn a lot. Study hard and go \\
& home, tomorrow. \\
CPM-Large: & 现在都是微积分,大学的东西比高中简单 \\
(Few-shot) & 多了,还有什么呢? \\
& Now it's all calculus. The things in college \\
& are much simpler than those of high school. \\
& What else? \\
CPM-Large: & 这是要逼死我吗?每道题都不会做啊  \\
(Supervised) & Is this going to kill me? I can’t answer \\
& any question. \\
\hline
\hline
Post: & 每天嘻嘻哈哈的笑着，只有自己才知道 \\
& 活的有多累。 \\
& I laugh every day, but only I know how \\
& tired I am. \\
\hline
CDial-GPT: & 我也是 \\
(Few-shot) & Me, too. \\
CDial-GPT: & 只有自己才知道活的有多累。 \\
(Supervised) & Only I know how tired I am. \\
CPM-Large: & 当你想要哭泣的时候,就去那些熟悉的地 \\
(Few-shot) & 方吧。 \\
& When you want to cry, go to those familiar \\
& places. \\
CPM-Large: & 真的不知道,生活怎么会这么累 \\
(Supervised) & I really don't know how my life could be \\
& so tiring. \\
\hline
\hline
\end{tabular}}
\caption{Examples of generated responses on STC.}
\label{tab:stccase}
\end{table}

\subsection{Dialogue Generation}

\textbf{Dataset}: We use Short-Text Conversation (STC) \cite{shang2015neural} as our benchmark dataset for dialogue generation, which consists of post-response pairs from Weibo. We adopt the same data split as the existing work \cite{wang2020large}. The amount of training / validation / test set is 4.4M / 20K / 20K, respectively. The average length of posts / responses is 20.6 / 15.4.

\textbf{Baseline}: We choose CDial-GPT \cite{wang2020large} as our baseline, which is the state-of-the-art pre-trained model for Chinese dialogue generation. We directly use the codes and the pre-trained model released by the original paper.

\textbf{Implementation Details}: In the supervised experiment, we utilize a similar hyper-parameter setting as pre-training and fine-tune CPM on the training set of STC. In the few-shot experiment which doesn't include the fine-tuning process, 
we follow the existing work \cite{radford2019language,brown2020language} to condition the language model on a context of 4 examples pairs of the format \textit{Context: sentence Response: sentence}. After a final prompt \textit{Context: sentence Response:}, we acquire the generation results with Top-$p$ sampling \cite{holtzman2019curious}, where $p$ is set to 0.9. The temperature of sampling is 0.9 in both few-shot and supervised experiments.

\textbf{Metrics}: Since BLEU is not a proper metric for dialogue generation, we use embedding-based metrics (including greedy matching, embedding average, and vector extrema) to evaluate the similarity between generated responses and references \cite{liu2016not}. For diversity, we choose the number and proportion of distinct n-grams \cite{li2016diversity,xing2017topic,ke2018generating} as our metric.

\textbf{Results}: We present the main results in the few-shot and supervised settings in Table \ref{tab:stcresult}. We can see that CPM outperforms CDial-GPT with a large margin in the few-shot experiment, showing the generalization ability of our model. As for the supervised experiment, our model still performs better, especially on the diversity metrics. Since fine-tuning large pre-trained models on the supervised downstream tasks is often challenging \cite{dodge2020fine,mosbach2020stability,lee2020mixout}, we leave how to further improve the performance in the supervised setting as future work. Some cases are provided in Table \ref{tab:stccase} to intuitively show the effectiveness of our model.

We also conduct experiments to show the few-shot performance of CPM with different parameter sizes in Table \ref{tab:stcdifferentparam}. As the number of parameters grows, CPM can generate more diverse responses with reasonable values on the embedding-based metrics.

\begin{table} [t]
\centering
\small
\setlength{\tabcolsep}{1.5mm}{
\begin{tabular}{lccccc}
\hline
& Average &  Dist-1 & Dist-2 \\
\hline
CPM-Small & \textbf{0.928} & 2,201 / 0.004 & 22,754 / 0.046   \\
CPM-Medium & 0.910 & 2,842 / 0.005 & 31,934 / 0.058   \\
CPM-Large & \textbf{0.928} & \textbf{3,229} / \textbf{0.007} & \textbf{68,008} / \textbf{0.154}   \\
\hline
\end{tabular}}
\caption{Results of CPM with different amounts of parameters on STC dataset in the few-shot setting.}
\label{tab:stcdifferentparam}
\end{table}

\begin{table}[htbp]
  \centering
  \small
    \begin{tabular}{l|cccccc}
    \hline
          & \multicolumn{2}{c}{\textcolor[rgb]{ .286,  .286,  .286}{Zhidao}} & \multicolumn{2}{c}{\textcolor[rgb]{ .286,  .286,  .286}{Search}} & \multicolumn{2}{c}{\textcolor[rgb]{ .286,  .286,  .286}{CMRC2018}} \\
    \cline{2-7}
          & \textcolor[rgb]{ .286,  .286,  .286}{F1} & \textcolor[rgb]{ .286,  .286,  .286}{EM} & \textcolor[rgb]{ .286,  .286,  .286}{F1} & \textcolor[rgb]{ .286,  .286,  .286}{EM} & \textcolor[rgb]{ .286,  .286,  .286}{F1} & \textcolor[rgb]{ .286,  .286,  .286}{EM} \\
    \hline
    \textcolor[rgb]{ .286,  .286,  .286}{s + zs} & \textcolor[rgb]{ .286,  .286,  .286}{4.01} & \textcolor[rgb]{ .286,  .286,  .286}{0.18} & \textcolor[rgb]{ .286,  .286,  .286}{4.15} & \textcolor[rgb]{ .286,  .286,  .286}{0.65} & \textcolor[rgb]{ .286,  .286,  .286}{6.03} & \textcolor[rgb]{ .286,  .286,  .286}{0.20} \\
    \textcolor[rgb]{ .286,  .286,  .286}{s + os} & \textcolor[rgb]{ .286,  .286,  .286}{4.75} & \textcolor[rgb]{ .286,  .286,  .286}{0.34} & \textcolor[rgb]{ .286,  .286,  .286}{4.45} & \textcolor[rgb]{ .286,  .286,  .286}{0.59} & \textcolor[rgb]{ .286,  .286,  .286}{6.14} & \textcolor[rgb]{ .286,  .286,  .286}{0.22} \\
    \textcolor[rgb]{ .286,  .286,  .286}{m + zs} & \textcolor[rgb]{ .286,  .286,  .286}{5.29} & \textcolor[rgb]{ .286,  .286,  .286}{0.29} & \textcolor[rgb]{ .286,  .286,  .286}{5.03} & \textcolor[rgb]{ .286,  .286,  .286}{0.61} & \textcolor[rgb]{ .286,  .286,  .286}{8.60} & \textcolor[rgb]{ .286,  .286,  .286}{0.53} \\
    \textcolor[rgb]{ .286,  .286,  .286}{m + os} & \textcolor[rgb]{ .286,  .286,  .286}{5.76} & \textcolor[rgb]{ .286,  .286,  .286}{0.47} & \textcolor[rgb]{ .286,  .286,  .286}{5.14} & \textcolor[rgb]{ .286,  .286,  .286}{0.55} & \textcolor[rgb]{ .286,  .286,  .286}{9.00} & \textcolor[rgb]{ .286,  .286,  .286}{0.75} \\
    \textcolor[rgb]{ .286,  .286,  .286}{l + zs} & \textcolor[rgb]{ .286,  .286,  .286}{5.18} & \textcolor[rgb]{ .286,  .286,  .286}{0.27} & \textcolor[rgb]{ .286,  .286,  .286}{5.08} & \textcolor[rgb]{ .286,  .286,  .286}{0.59} & \textcolor[rgb]{ .286,  .286,  .286}{13.37} & \textcolor[rgb]{ .286,  .286,  .286}{1.31} \\
    \textcolor[rgb]{ .286,  .286,  .286}{l + os} & \textcolor[rgb]{ .286,  .286,  .286}{\textbf{6.08}} & \textcolor[rgb]{ .286,  .286,  .286}{\textbf{0.56}} &  \textcolor[rgb]{ .286,  .286,  .286}{\textbf{5.19}}     &   \textcolor[rgb]{ .286,  .286,  .286}{\textbf{0.68}}    & \textcolor[rgb]{ .286,  .286,  .286}{\textbf{16.56}} & \textcolor[rgb]{ .286,  .286,  .286}{\textbf{3.73}} \\
    \hline
    \end{tabular}%
    \caption{Zero-shot (zs) and one-shot (os) results on Question Answering (QA) datasets, including DuReader (Zhidao and Search) and CMRC2018, we did experiments on models with three different sizes: small (s), medium (m) and large(l).}
  \label{tab:qa}%
\end{table}%

\subsection{Question Answering}
\textbf{Dataset}: We adopt CMRC2018~\cite{CMRC2018} and DuReader~\cite{DuReader} as our benchmark for Question Answering (QA).  CMRC2018 requires the model to extract an answer span from a Wikipedia passage for the given question, which is similar to SQuAD~\cite{SQuAD}. DuReader consists of questions from real-world user logs from Baidu Search and Baidu Zhidao. The answers in DuReader are manifold, such as an entity or a description. We treat DuReader as an extractive QA task and thus ignore those instances with yes-or-no answers during evaluation.

\textbf{Implementation Details}: We evaluate CPM on zero-shot (zs) and one-shot (os) setting and report F1 score (F1) and Exact Match (EM) for both CMRC2018 and DuReader. For the zero-shot setting, we concatenate the passage and question as input to CPM, and CPM is then required to generate an answer according to the observed (passage, question) pair. For the one-shot setting, we randomly select a ground truth triple (passage, question, answer) in the training set and insert it to the front of the instance to be treated as a hint for CPM to generate the answer.

\textbf{Results}:  As shown in Table \ref{tab:qa}, we perform the experiments on three datasets and compare models with different sizes: small (s), medium (m) and large (l). From the table, we can see that with the size growing, CPM is performing better. Among all the models, large is always the best. And, the results in the one-shot setting are better than those in the zero-shot setting. We guess CPM is able to imitate the format in previous sequences and organize the language accordingly. We also analyze the generated answer and find that CPM prefers to generate long and repetitive sentences instead of a short and precise one, which results in low scores. We believe it is worth exploring how to make CPM generate brief and proper answers in the future. In general, CPM does not achieve very high scores in either benchmark. We guess it's related to the format of the pre-training data.

\begin{table*}[t]
\begin{center} 
\scalebox{0.9}{ 
\setlength{\tabcolsep}{1pt}{
\begin{tabular}{l|ccc|ccc}
\hline
  & \multicolumn{3}{c|}{\textbf{$N=2$}} & \multicolumn{3}{c}{\textbf{$N=4$}}  \\ 
  \hline
  \multicolumn{1}{c|}{CPM} & \multicolumn{1}{c}{Small} & \multicolumn{1}{c}{Medium} & \multicolumn{1}{c|}{Large} & \multicolumn{1}{c}{Small} & \multicolumn{1}{c}{Medium} & \multicolumn{1}{c}{Large} \\
\hline
主要工艺~(Main Process) & \multicolumn{1}{c}{0.500} & \multicolumn{1}{c}{0.500} & \multicolumn{1}{c|}{\textbf{0.700}} & \multicolumn{1}{c}{0.400} & \multicolumn{1}{c}{0.200} & \multicolumn{1}{c}{\textbf{0.400}}  \\
释义~(Explanation) & \multicolumn{1}{c}{0.000} & \multicolumn{1}{c}{0.000} & \multicolumn{1}{c|}{\textbf{0.071}} & \multicolumn{1}{c}{0.000} & \multicolumn{1}{c}{0.000} & \multicolumn{1}{c}{\textbf{0.075}}  \\
商品品牌~(Brand) & \multicolumn{1}{c}{0.098} & \multicolumn{1}{c}{0.033} & \multicolumn{1}{c|}{\textbf{0.483}} & \multicolumn{1}{c}{0.183} & \multicolumn{1}{c}{0.050} & \multicolumn{1}{c}{\textbf{0.450}}  \\
学科~(Subject) & \multicolumn{1}{c}{0.000} & \multicolumn{1}{c}{0.025} & \multicolumn{1}{c|}{\textbf{0.124}} & \multicolumn{1}{c}{0.059} & \multicolumn{1}{c}{0.053} & \multicolumn{1}{c}{\textbf{0.108}}  \\
全名~(Full Name) & \multicolumn{1}{c}{0.035} & \multicolumn{1}{c}{0.010} & \multicolumn{1}{c|}{\textbf{0.108}} & \multicolumn{1}{c}{0.000} & \multicolumn{1}{c}{0.014} & \multicolumn{1}{c}{\textbf{0.122}}  \\
涉及领域~(Related Field)  & \multicolumn{1}{c}{0.042} & \multicolumn{1}{c}{0.065} & \multicolumn{1}{c|}{\textbf{0.104}} & \multicolumn{1}{c}{0.063} & \multicolumn{1}{c}{0.037} & \multicolumn{1}{c}{\textbf{0.125}}  \\
主要作物~(Main Crop) & \multicolumn{1}{c}{0.000} & \multicolumn{1}{c}{\textbf{0.150}} & \multicolumn{1}{c|}{0.050} & \multicolumn{1}{c}{0.100} & \multicolumn{1}{c}{\textbf{0.150}} & \multicolumn{1}{c}{0.100}  \\
所在国家~(In Country) & \multicolumn{1}{c}{0.033} & \multicolumn{1}{c}{0.033} & \multicolumn{1}{c|}{0.033} & \multicolumn{1}{c}{\textbf{0.050}} & \multicolumn{1}{c}{0.000} & \multicolumn{1}{c}{\textbf{0.050}}  \\
病原类型~(Pathogen Type) & \multicolumn{1}{c}{0.250} & \multicolumn{1}{c}{0.220} & \multicolumn{1}{c|}{\textbf{0.370}} & \multicolumn{1}{c}{0.200} & \multicolumn{1}{c}{0.300} & \multicolumn{1}{c}{\textbf{0.340}}  \\
首任总统~(The First President)  & \multicolumn{1}{c}{0.000} & \multicolumn{1}{c}{0.000} & \multicolumn{1}{c|}{0.000} & \multicolumn{1}{c}{\textbf{0.016}} & \multicolumn{1}{c}{0.009} & \multicolumn{1}{c}{0.014}  \\
\hline
\end{tabular}}} 
\end{center} 
\caption{BLEU-1 results of CPM with different amounts of parameters on XLORE dataset in the few-shot setting.}
\label{tab:entitydifferentparam}
\end{table*}

\begin{table} [t]
\centering
\footnotesize
\setlength{\tabcolsep}{0.8mm}{
\begin{tabular}{ll}
\hline
Relation: & 首都~(Capital) \\
\hline
Prompt: & 美国\ 首都\ 华盛顿 \\
& America\ Capital\ Washington \\
& 中国\ 首都\ 北京 \\
& China\ Capital\ Beijing \\
& 日本\ 首都 \\
& Japan\ Capital \\
CPM: & 东京  \\
& Tokyo \\
\hline
\hline
Relation: & 主要工艺~(Main Process) \\
\hline
Prompt: & 酱焖辣椒\ 主要工艺\ 焖 \\
& (Sauce Braised Chili)\ (Main Process)\ Stew\\
& 当归鸭肉煲\ 主要工艺\ 煲 \\
& (Duck with Angelica)\ (Main Process)\ Boil \\
& 韭菜煎蛋饼\ 主要工艺 \\
& (Leek Omelette)\ (Main Process) \\
CPM: & 煎  \\
& Fried  \\
\hline
\hline
Relation: & 学科~(Subject) \\
\hline
Prompt: & 恒星级黑洞\ 学科\  宇宙论\\
& (Stellar Black Hole)\ Subject\  Cosmology\\
& 品类需求强度\ 学科\  品牌经济学 \\
& (Category Demand Intensity)\ Subject\ Economics \\
& 大地构造学\ 学科\  \\
& (Tectonic Geology)\ Subject\  \\
CPM: & 地质学 \\
& Geology \\
\hline

\end{tabular}}
\caption{Examples of generated entities on XLORE with CPM-Large.}
\label{tab:XLOREcase}
\end{table}

\subsection{Entity Generation}
\textbf{Dataset}: We use XLORE, which includes 446,236 relations and 16,284,901 entities, as our benchmark dataset for entity generation. These relations and entities are from Wikipedia and Baidu Baike. 

\textbf{Implementation Details}: We evaluate CPM on the few-shot setting with different amounts of parameters and report BLEU-1 results. In detail, we randomly select triples~(head entity, relation, tail entity)~by the same relations from XLORE and combine $N$ triples and an incomplete triple~(head entity, relation) into a prompt. Then, given the prompt, the models need to predict the corresponding tail entity.

\textbf{Results}:  
We present the results in Table~\ref{tab:entitydifferentparam}. As we can see from the table, CPM-large achieves the best performance among these three models. Surprisingly, given a prompt with two triples, CPM can achieve comparable results to that with four triples. It indicates that CPM can imitate the format and probe factual knowledge to generate a proper tail entity in the extreme few-shot scenarios. We also provide some cases in Table~\ref{tab:XLOREcase} to demonstrate the ability of CPM.

\section{Future Work}

In the future, we will further explore the power of large-scale pre-trained models on Chinese by adding more training data and increasing the model size. Due to the extremely expensive cost of pre-training, we will try to optimize the training framework, such as the data-transfer scheme between different nodes, to accelerate the process. There are some previous works including LAMB~\cite{You2020Large} and DeepSpeed~\cite{rasley2020deepspeed}. Besides, it is important to reduce the model size by model compression~\cite{sanh2019distilbert,jiao2019tinybert,zhang2020know}.

Meanwhile, we will also include diverse data to enhance model performance. For text data, we will add a multi-lingual corpus to train a large-scale Chinese-centered multi-lingual language model. For structured data such as knowledge graphs, which is important for PLMs~\cite{DBLP:conf/emnlp/PetersNLSJSS19,DBLP:conf/iclr/XiongDWS20,su2020contextual}, we will explore new learning algorithms to train a joint model, which can learn from both texts and knowledge graphs for better general intelligence.

\section*{Acknowledgments}
Thanks to the Beijing Academy of Artiicial Intelligence (BAAI) for providing the computing resources and web services of this work. In addition, we would like to thank NetEase Inc., zhihu.com, and aminer.cn for the support in collecting the Chinese corpus.

\section*{Disclaimer of Warranties}
The text generated by CPM is automatically generated by a neural network model trained on a large number of texts, which does not represent our official attitudes and preferences. The text generated by CPM is only used for technical and scientific purposes. If it infringes on your rights and interests or violates social morality, please do not propagate it, but contact us and we will deal with it promptly.

\normalem
\bibliography{acl2020}
\bibliographystyle{acl_natbib}

\appendix

\section{Contributions}

\noindent \textbf{Zhengyan Zhang, Xu Han, and Hao Zhou}
implemented the large-scale models and model-parallel strategies.

\vbox{}

\noindent \textbf{Huanqi Cao, Shengqi Chen, Daixuan Li, and Zhenbo Sun}
built the training infrastructure.

\vbox{}

\noindent \textbf{Pei Ke, Deming Ye, Jian Guan, Fanchao Qi, and Xiaozhi Wang}
collected, filtered, deduplicated the training data.

\vbox{}

\noindent \textbf{Zhengyan Zhang, Pei Ke, Yuxian Gu, Deming Ye, Yujia Qin, Yusheng Su, and Haozhe Ji}
implemented the downstream tasks and the software framework for supporting them.

\vbox{}

\noindent \textbf{Hao Zhou, Guoyang Zeng, Xu Han, and Yanan Zheng}
implemented the demos of language generation and knowledge retrieval using our CPM.

\vbox{}

\noindent \textbf{Guoyang Zeng}
conducted the human evaluations of the model.

\vbox{}

\noindent \textbf{Hao Zhou, Zhengyan Zhang, Pei Ke, Yuxian Gu, Deming Ye, Yujia Qin, and Yusheng Su}
wrote the paper.

\vbox{}

\noindent \textbf{Zhiyuan Liu, Minlie Huang, and Wentao Han}
designed and led the research.

\vbox{}

\noindent \textbf{Jie Tang, Juanzi Li, Xiaoyan Zhu, Maosong Sun}
provided valuable advices to the research.

\end{CJK*}
\end{document}